\documentclass[sigconf]{acmart}

\usepackage{bm}
\usepackage{multirow}
\usepackage{xspace}
\usepackage[tight,footnotesize]{subfigure}
\usepackage{booktabs}
\usepackage{threeparttable}
\usepackage{enumitem}
\usepackage{color}
\usepackage{algpseudocode}
\usepackage{amsmath}
\usepackage[linesnumbered,ruled]{algorithm2e}
\newtheorem{myDef}{Definition}

\newtheorem{mytheo}{Theorem}
\newtheorem{myprof}{Proof}
\newcommand{\paratitle}[1]{\vspace{1.5ex}\noindent\textbf{#1}}

\newcommand{\eg}{\emph{e.g.,}\xspace}

\newcommand{\ignore}[1]{}

\renewcommand\footnotetextcopyrightpermission[1]{} %

\author{Zekun Tong}
\affiliation{%
 \institution{National University of Singapore}
}
\email{zekuntong@u.nus.edu}

\author{Yuxuan Liang}
\affiliation{%
 \institution{National University of Singapore}
}
\email{yuxliang@outlook.com}

\author{Changsheng Sun}
\affiliation{%
 \institution{National University of Singapore}
}
\email{changsheng_sun@outlook.com}

\author{David S. Rosenblum}
\affiliation{%
 \institution{National University of Singapore}
}
\email{david@comp.nus.edu.sg}

\author{Andrew Lim}
\authornote{Corresponding author}
\affiliation{%
 \institution{National University of Singapore}
}
\email{isealim@nus.edu.sg}

\begin{document}
\title[Directed Graph Convolutional Networks]{Directed Graph Convolutional Network}

\begin{abstract}
Graph Convolutional Networks (GCNs) have been widely used due to their outstanding performance in processing graph-structured data. However, the undirected graphs limit their application scope. In this paper, we extend spectral-based graph convolution to directed graphs by using first- and second-order proximity, which can not only retain the connection properties of the directed graph, but also expand the receptive field of the convolution operation. A new GCN model, called DGCN, is then designed to learn representations on the directed graph, leveraging both the first- and second-order proximity information. We empirically show the fact that GCNs working only with DGCNs can encode more useful information from graph and help achieve better performance when generalized to other models. Moreover, extensive experiments on citation networks and co-purchase datasets demonstrate the superiority of our model against the state-of-the-art methods.
\end{abstract}



\settopmatter{printacmref=false}

\maketitle

\section{Introduction}
\label{intro}

Graph structure is a common form of data. Graphs have a very strong ability to represent complex structures and can easily express entities and their relationships. Graph Convolutional Networks (GCNs)~\cite{hammond2011wavelets,defferrard2016convolutional,kipf2016semi,hamilton2017inductive,velivckovic2017graph,xu2018powerful} are a CNNs variant on graphs and effectively learns underlying pairwise relations among data vertices. We can divide GCNs into two categories: spectral-based~\cite{kipf2016semi,li2018adaptive,defferrard2016convolutional} and spatial-based~\cite{hamilton2017inductive,velivckovic2017graph}. A representative structure of spectral-based GCNs~\cite{kipf2016semi} has multiple layers that stack first-order spectral filters and learn graph representations using a nonlinear activation function, while spatial-based approaches design neighborhood features aggregating schemes to achieve graph convolutions~\cite{wu2019comprehensive,ying2019gnnexplainer}. Various GCNs have significant improvements in benchmark datasets~\cite{kipf2016semi,hamilton2017inductive,velivckovic2017graph}. Such breakthroughs have promoted the exploration of variant networks: the Attention-based Graph Neural Network(AGNN)~\cite{thekumparampil2018attention} uses the attention mechanism  to replace the propagation layers in order to learn dynamic neighborhood information; the FastGCN~\cite{chen2012svdfeature} interprets graph convolutions as integral transforms of embedding functions while using batch training to speed up. The positive effects of different GCN variants greatly promote their use in various task fields, including but not limited to social networks~\cite{chen2012svdfeature}, quantum chemistry~\cite{liao2019lanczosnet}, text classification~\cite{yao2019graph} and image recognition ~\cite{wang2018zero}.

One of the main reasons that a GCN can achieve such good results on a graph is that makes full use of the structure of the graph. It captures rich information from the neighborhood of object nodes through the links, instead of being limited to a specific distance range. General GCN models provide a neighborhood aggregation scheme for each node to gain a representation vector, and then learn a mapping function to predict the node attributes~\cite{hou2020measuring}. Since spatial-based methods need to traverse surrounding nodes when aggregating features, they usually add significant overhead to computation and memory usage~\cite{wu2019simplifying}. On the contrary, spectral-based methods use matrix multiplication instead of traversal search, which greatly improves the training speed. Thus, in this paper, we focus mainly on the spectral-based method.


Although the above methods have achieved improvement in many aspects, there are two major shortcomings with existing spectral-based GCN methods.

First, spectral-based GCNs are limited to apply to undirected graphs~\cite{wu2019comprehensive}. For directed graphs, the only way is to relax the graph structure from a directed to an undirected graph by symmetrizing the adjacency matrix. In this way we can get the semi-definite Laplacian matrix, but at the same time we also lose the unique structure of the directed graph. For example, in a citation graph, later published articles can cite former published articles, but the reverse is not true. This is a unique time series relationship. If we transform it into an undirected graph, we lose this part of the information. Although we can represent the original directed graph in another form, such a temporal graph learned by combination of Recurrent Neural Networks (RNNs) and GCNs ~\cite{pareja2019evolvegcn}, we still want to dig more structural features from the original without adding extra components.

Second, in most existing spectral-based GCN models, during each convolution operation, only 1-hop node features are taken into account (using the same adjacency matrix), which means they only capture the \emph{first-order} information between vertices. It is natural to consider direct links when extracting local features, but this is not always the case. In some real-world graph data, many legitimate relationships may not be encoded via first-order links~\cite{tang2015line}. For instance, people in social networking communities share common interests, but they don not necessarily contact each other. In other words, the features we get by \emph{first-order} proximity are likely to be insufficient. Although we can obtain more information by stacking multiple layers of GCNs, multi-layer networks will introduce more trainable parameters, making them prone to overfitting when the label rate is small or needing extra labeled information, which is not label efficient~\cite{li2019label}. Therefore, we need a more efficient feature extraction method.

\begin{figure}[t]
  
  \centering
  \setlength{\belowcaptionskip}{-0.9cm} 
  \includegraphics[width=5cm]{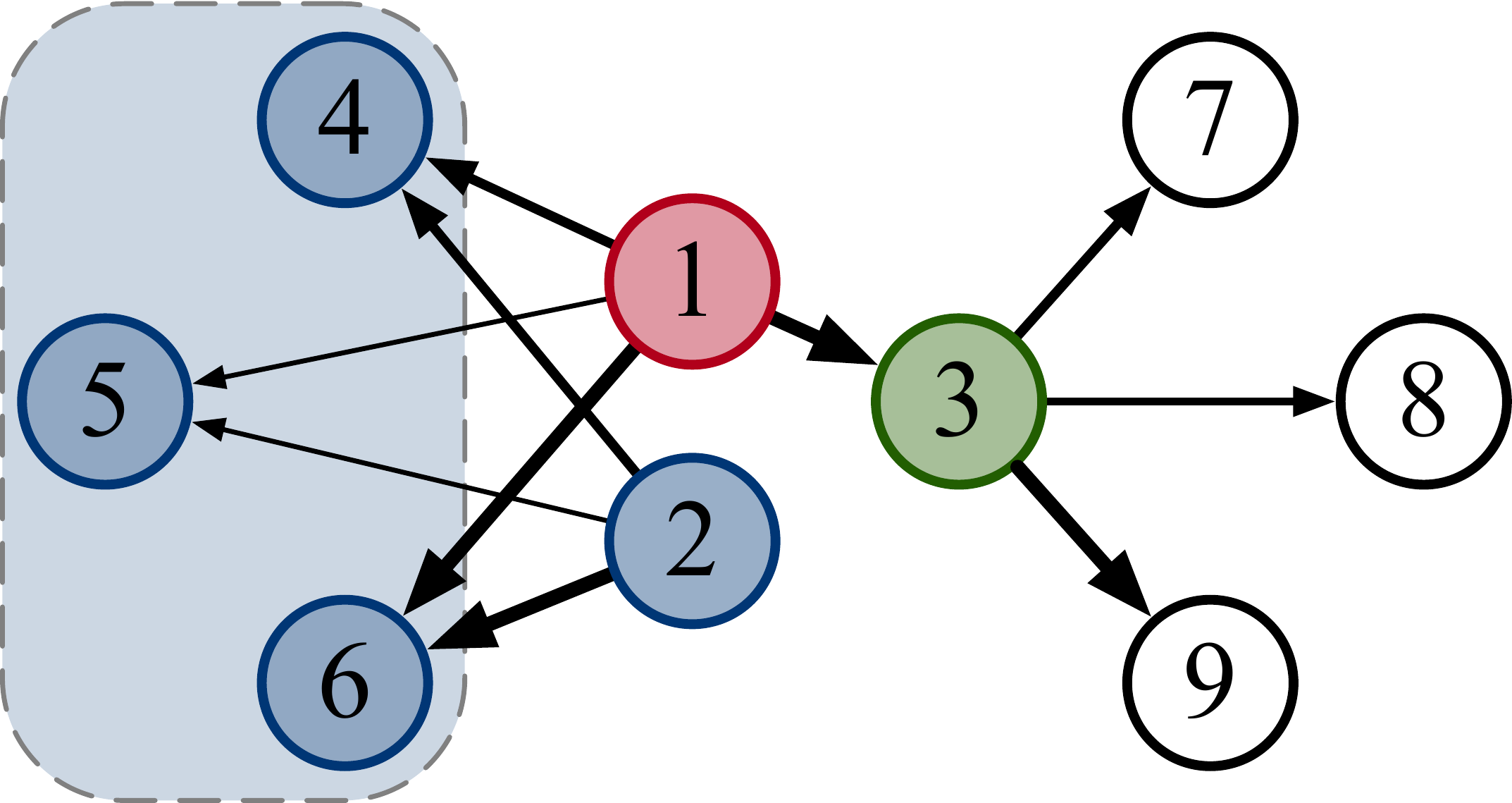}\\
  \caption{A simple weighted directed graph example. Line width indicates the weight of the edges. The node $v_1$ has \emph{first-order} proximity with $v_3$. It also has \emph{second-order} proximity with $v_2$, because they have shared neighbors $\{v_4,v_5,v_6\}$. Both $v_2$ and $v_3$'s features should be considered when aggregating $v_1$'s feature.}\label{second_order}
\end{figure}

To address these issues, we leverage \emph{second-order} proximity between vertices as a complement to existing methods, which inspire from the $hub$ and $authority$ web model~\cite{kleinberg1999authoritative,zhou2005semi}. By using \emph{second-order} proximity, the directional features of the directed graph can be retained. Additionally, the receptive field of the graph convolution can be expanded to extract more features. Different from \emph{first-order} proximity, judging whether two nodes have \emph{second-order} proximity does not require nodes to have paths between them, as long as they share common neighbors, which can be considered as \emph{second-order} proximity. In other words, nodes with more shared neighbors have stronger \emph{second-order} proximity in the graph. This general notion also appears in sociology~\cite{granovetter1983strength}, psychology~\cite{reich2012friending} and daily life: people who have a lot of common friends are more likely to be friends. A simple example is shown in Figure \ref{second_order}. When considering the neighborhood feature aggregation of $v_1$ in the $layer_1$, we need to consider $v_3$, because it has a \emph{first-order} connection with $v_1$, but we also need to aggregate $v_2$'s features, due to the high \emph{second-order} similarity with $v_1$ in sharing three common neighbors.

In this paper, we present a new spectral-based model on directed graphs, \underline{\textbf{D}}irected \underline{\textbf{G}}raph \underline{\textbf{C}}onvolutional \underline{\textbf{N}}etworks (\textbf{DGCN}), which utilizes \emph{first \& second-order} proximity to extract graph information. We not only consider basic \emph{first-order} proximity to obtain neighborhood information, but also take \emph{second-order} proximity into account, which is a complement for sparsity of \emph{first-order} information in real-world data. What's more, we verify this efficiency by extending \emph{Feature and Label Smoothness} measurements ~\cite{hou2020measuring} to our application scope. Through experiments, we empirically show that our model exhibits superior performance against baselines while the number of parameters and computational consumption are significantly reduced.


In summary, our study has the following contributions:

\begin{enumerate}[leftmargin=*]
  \item We present a novel graph convolutional networks called the \textbf{"DGCN"}, which can be applied to the directed graphs by utilizing \emph{first- and second-order} proximity. To our knowledge, this is the first-ever attempt that enables the spectral-based GCNs to generalize to directed graphs.
  \item We define \emph{first- and second-order} proximity on the directed graph, which are designed for expanding the convolution operation receptive field, extracting and leveraging graph information. Meanwhile, we empirically show that this method has both feature and label efficiency, while also demonstrating powerful generalization capability. 
  \item We experiment with semi-supervised classification tasks on various real-world datasets, which validates the effectiveness of \emph{first- and second-order} proximity and the improvements obtained by DGCNs over other models. We will release our code for public use soon.
\end{enumerate}

\section{Preliminaries}

In this section, we first clarify the terminology and preliminary knowledge of our study, and then present our task in detail. Particularly, we use bold capital letters (\eg $\mathbf{X}$) and bold lowercase letters (\eg $\mathbf{x}$) to denote matrices and vectors, respectively. We use non-bold letters (\eg $x$) to represent scalars and Greek letters (\eg $\lambda$) as parameters.

\subsection{Formulation}
\label{for}

\begin{myDef}
\textbf{Directed Graph}~\cite{poignard2018spectra}. A general graph has $n$ vertices or nodes is define as ${\mathcal{G}} = (\mathcal{V}, \mathcal{E})$, where $\mathcal{V} = (v_1,...,v_n)$ is vertex set and $\mathcal{E} \subset \{1,..,n\}\times \{1,...,n\} $ is edge set. Each edge $e\in \mathcal{E} $ is an ordered pair $e = (i,j)$ between vertex $v_i$ and $v_j$. If any of its edges $e$ is associated with a weight $W_{i,j}>0$, the graph is weighted. When pair $(i,j) \notin \mathcal{E}$, we set $W_{i,j} = 0$. A graph is directed when has $(u,v) \not\equiv (v,u)$ and $w_{i,j} \not\equiv  w_{j,i}$.  
\end{myDef}

Here, we consider the undirected graph as a directed graph has two directed edges with opposite directions and equal weights, and binary graph as a weighted graph which edge weight values only take from 0 or 1. Besides, we only consider non-negative weights.


Furthermore, to measure our model's ability to extract surrounding information, we define two indicators\cite{hou2020measuring,wang2019knowledge} on the directed graph: \textbf{Feature Smoothness}, which is used to evaluate how much surrounding information we can obtain and \textbf{Label Smoothness} for evaluating how useful the obtained information is.

\begin{myDef}
\textbf{First \& second-order edges in directed graph}. Given a directed graph $\mathcal{G} = (\mathcal{V}, \mathcal{E})$. For an order pair $(i,j)$, where $i$ and $j$ $\in \mathcal{V} $. If $(i,j)\in \mathcal{E}$, we say that order pair $(i,j)$ is first-order edge. If it exists any vertex $k \in \mathcal{V}$ that satisfies order pairs $(k,i)~and~(k,j) \in \mathcal{E}~or~(i,k)~and~(j,k ) \in \mathcal{E}$, we define order pair $(i,j)$ as second-order edge. The edge set has both first \& second-order edge of $\mathcal{G}$denoted by $\mathcal{E}'$.
\end{myDef}

\begin{myDef} \textbf{Feature Smoothness}. We define the Feature Smoothness $\lambda_{f}$ over normalized node feature space $\mathcal{X} = [0,1]^d $ as follows,	

\begin{equation}
	\lambda_{f}=\frac{\left\| \sum_{i\in \mathcal{V'} } \left ( \sum_{e_{i, j} \in \mathcal{E'}} \left(x_{i}-x_{j}\right)^{2}\right)\right\|_{1}}{|\mathcal{E}'| \cdot d},
	\end{equation}
where ${||} \cdot {||}_1$ is  the Manhattan norm, $d$ is the node feature dimension and $x_i$ is feature of vertex $i$.
\end{myDef}

\begin{myDef} \textbf{Label Smoothness}. For the node classification task, we want to determine how useful the information obtained from the surrounding nodes is. We believe that the information obtained is valid when the current node label is consistent with the surrounding node labels, otherwise it is invalid. Based on this, we define Label Smoothness $\lambda_{l}$ as:

	\begin{equation}
	\lambda_{l}=\frac {\sum_{e_{i, j} \in \mathcal{E'}} \left(\mathbb{I}\left(v_{i} \simeq v_{j}\right)\right)}{|\mathcal{E'}|},
	\end{equation}
where $\mathbb{I}(\cdot)$ is an indictor function and we define $v_{i} \simeq v_{j} $ if the label of vertex $i$ and $j$ are the same.
\end{myDef}


\subsection{Problem Statement}

After giving the above formulation, we are ready to define our task.

\begin{myDef}
\textbf{Semi-Supervised Node Classification}\cite{abu2018n}. Given a graph $\mathcal{G} = (\mathcal{V}, \mathcal{E})$ with adjacency matrix $A$, and node feature matrix $\mathbf{X} \in \mathbb {R}^{N \times C}$, where $N=|\mathcal{V}|$ is the number of nodes and $C$ is the feature dimension. Given a subset of nodes $\mathcal{V}_L \subset \mathcal{V}$, where nodes in $\mathcal{V}_L$ have observed labels and generally $|\mathcal{V}_L|<<|\mathcal{V}|$. The task is using the labeled subset $\mathcal{V}_L$, node feature matrix $X$ and adjacency matrix $A$ predict the unknown label in $\mathcal{V}_{UL} = \mathcal{V} \setminus \mathcal{V}_L$.

\end{myDef}

\section{Undirected Graph Convolution}
\label{un}
In this section, we will go through the spectral graph convolutions defined on the undirected graph.


We have the undirected graph Laplacian $\mathbf{L'} = \mathbf{D'}-\mathbf{A'}$, $\mathbf{A'}$ is the adjacency matrix of the graph, $\mathbf{D'}$ is a diagonal matrix of node degree, $D'_{ii} = \sum_j({A'}_{i,j})$. $\mathbf{L'}$ can be factored as $\mathbf{L'}= \mathbf{I} -\mathbf{D'}^{-\frac{1}{2}} \mathbf{A'} \mathbf{D'}^{-\frac{1}{2}} = \mathbf{U'} \mathbf{\Lambda} \mathbf{U'}^{T}$, where $\mathbf{I}$ is identity matrix, $\mathbf{U'}$ is the matrix of eigenvectors and $\mathbf{\Lambda}$ is the diagonal matrix of eigenvalues. The spectral convolutions on graph is defined as the multiplication of node feature vector $\mathbf{x} \in \mathbb{R}^N$ with a filter $\mathbf{g} \in \mathbb{R}^N $ in the Fourier domain:
\begin{equation}
\label{eq0}
	x * g =\mathbf{U'}\left(\mathbf{U'}^{T} x \odot \mathbf{U'}^{T} g \right)=\mathbf{U'} g_{\theta} \mathbf{U'}^{T} x,
\end{equation}
where $g_{\theta} = diag(\mathbf{U}^{T}\mathbf{x})$, $*$ represents convolution operation and $\odot$ is the element-wise Hadamard product.

Graph convolutions can be further approximated by $k^{th}$ Chebyshev polynomials to reduce computation-consuming: 


\begin{equation}
\label{cheb}
	\mathbf{x} * \mathbf{g}_{\theta} \approx \sum_{i=1}^{K} \theta_{i} T_{i}(\tilde{L'}) x,
\end{equation}
where $\mathbf{\tilde{L'}}=2 \mathbf{L'} / \lambda_{\max}-\mathbf{I}$, $\lambda_{\max}$ denotes the largest eigenvalue of $L'$ and $T_{k}(x)=2 x T_{k-1}(x)-T_{k-2}(x)$ with $T_{0}(x)=1$ and $T_{1}(x)=x$. The K-polynomial filter $\mathbf{g}_{\theta}$ shows its good localization in the vertex domain through integrating the node features within the K-hop neighborhood\cite{zhang2019graph}. Our model obtains node features in a different way, which will be explained in the Section \ref{diss}.

Kipf et al. employ Graph Convolutional Networks (GCNs)\cite{kipf2016semi}, which is a first-order approximation of ChebNet. They assume $k=1$, $\lambda_{max} = 2$, $\theta_{0}=2\theta$ and $\theta_{1} = -\theta$ in Equation \ref{cheb} to simplify the convolution process and get the following GCN convolution operation:

\begin{equation}
\label{kipf_complex}
	\mathbf{x} * \mathbf{g}_{\theta} \approx \theta\left(\mathbf{I}+\mathbf{D'}^{-\frac{1}{2}} \mathbf{A'} \mathbf{D'}^{-\frac{1}{2}}\right) \mathbf{x}.
\end{equation}

They also use renormalization trick converting $\mathbf{I}+\mathbf{D'}^{-1 / 2} \mathbf{A'} \mathbf{D'}^{-1 / 2} $ to $\mathbf{\tilde{D'}}^{-1 / 2} \mathbf{\tilde{A'}} \mathbf{\tilde{D'}}^{-1 / 2}$, where $\mathbf{\tilde{A'}}=\mathbf{A'}+\mathbf{I}$ and $\tilde{D'_{ii}} = \sum_j(\tilde{{A'}}_{i,j})$, to alleviate numerical instabilities and exploding/vanishing gradients problems. The final graph convolution layer is defined as follows:
\begin{equation}
	\mathbf{Z'} = \left(\mathbf{\tilde{D'}}^{-\frac{1}{2}} \mathbf{\tilde{A'}} \mathbf{\tilde{D'}}^{-\frac{1}{2}}\right) \mathbf{X} \Theta.
\label{kipf}
\end{equation}
Here, $\mathbf{X}\in \mathbb{R}^{N \times C}$ is the C-dimensional node feature vector, $\Theta \in \mathbb{R}^{C \times F} $ is the filter parameters matrix and $\mathbf{Z'} \in \mathbb{R}^{N \times F}$ is the convolved result with $F$ output dimension.

However, these derivations are based on the premise that Laplacian matrices are undirected graphs. The way they use to deal with directed graph is to relax it into an undirected graph, thereby constructing a symmetric Laplacian matrix. Although the relaxed matrix can be used for spectral convolution, it is not able to represent the actual structure of the directed graph. For instance, as we mention in the Section \ref{intro}, there is a time limit for citations: previously published papers cannot cite later ones. If the citation network is relaxed to an undirected graph, this restriction will no longer exist. 

In order to solve this problem, we propose a spectral-base GCN model for directed graph that leverages First- and Second-order Proximity in the following sections.

\section{The New Model: DGCN}

In this section, we present our spectral-based GCN model $f(\mathbf{X},\mathbf{A})$ for directed graphs that leverages the First- and Second-Order Proximity, called \textbf{DGCN}. We provide the mathematical motivation of directed graph convolution and consider a multi-layer Graph Convolutional Network which has the following layer-wise propagation rule, where 


\begin{eqnarray*}
	\mathbf{\hat{A}_F} & = & \mathbf{\tilde{D}_F}^{-\frac{1}{2}} ~~\mathbf{\tilde{A}_F} ~~\mathbf{\tilde{D}_F}^{-\frac{1}{2}} \\
	\mathbf{\hat{A}_{S_{in}}} & =&\mathbf{\tilde{D}_{S_{in}}}^{-\frac{1}{2}} ~~\mathbf{\tilde{A}_{S_{in}}} ~~\mathbf{\tilde{D}_{S_{in}}}^{-\frac{1}{2}}\\
	\mathbf{\hat{A}_{S_{out}}} & = &\mathbf{\tilde{D}_{S_{out}}}^{-\frac{1}{2}} ~~\mathbf{\tilde{A}_{S_{out}}} ~~\mathbf{\tilde{D}_{S_{out}}}^{-\frac{1}{2}}, 
\end{eqnarray*}
and 
\begin{eqnarray}
\label{eq1}
	\mathbf{H}^{(l+1)}=\Gamma \begin{pmatrix}
\sigma(\mathbf{\hat{A}_F}\mathbf{H}^{(l)}{\Theta}^{(l)}),\sigma(\mathbf{\hat{A}_{S_{in}}}\mathbf{H}^{(l)} {\Theta}^{(l)}),\sigma(\mathbf{\hat{A}_{S_{out}}} \mathbf{H}^{(l)}{\Theta}^{(l)})
\end{pmatrix}.
\end{eqnarray}
Here, $\mathbf{\tilde{A}_F}$ is the normalized First-order Proximity matrix with self-loop and $\mathbf{\tilde{A}_{S_{in}}},\mathbf{\tilde{A}_{S_{out}}}$ are the normalized Second-order Proximity matrix with self-loop, which are defined in Section \ref{matrix}. $\Gamma(\cdot)$ is a fusion function combines the proximity matrices together defined in Section \ref{dgc}. ${\Theta}^{(l)}$ is a shared trainable weight matrix and $\sigma(\cdot)$ is an activation function. $\mathbf{H}^{(l)}\in \mathbb{R}^{N \times C}$ is the matrix of activation in the $l^{th}$ layer and $\mathbf{H}^{(0)} = \mathbf{X}$. Finally, we present our model using directed graph convolution for semi-supervised learning in detail.

\begin{figure}[!h]
  \centering
  \includegraphics[width=6.8cm]{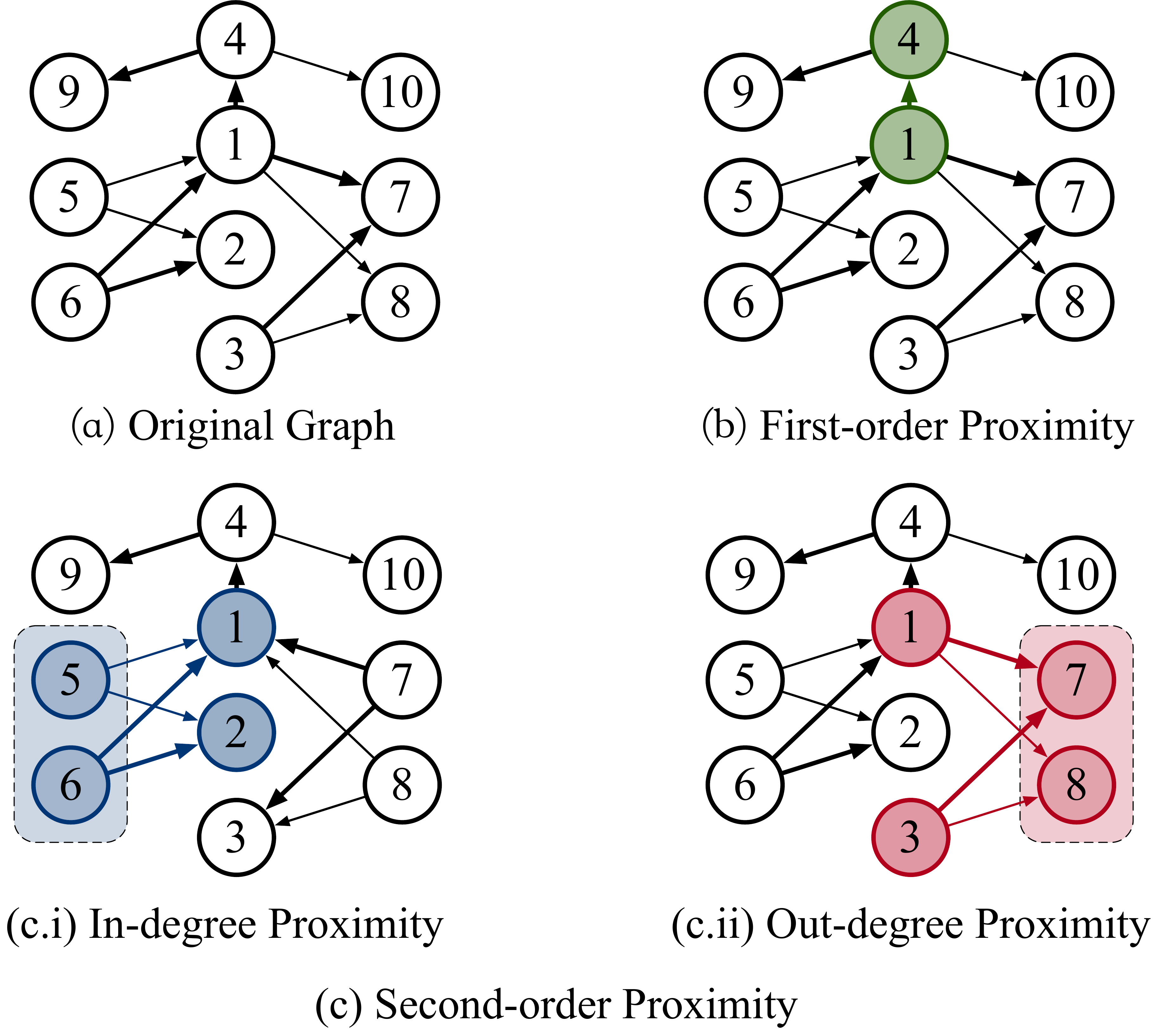}\\
  \setlength{\belowcaptionskip}{-0.5cm} 
  \caption{First- and second-order proximity examples in a weighted directed graph.}
  \label{origin_first_in_out}
\end{figure}

\begin{figure*}[t]
  \centering
  \includegraphics[width=16.5cm]{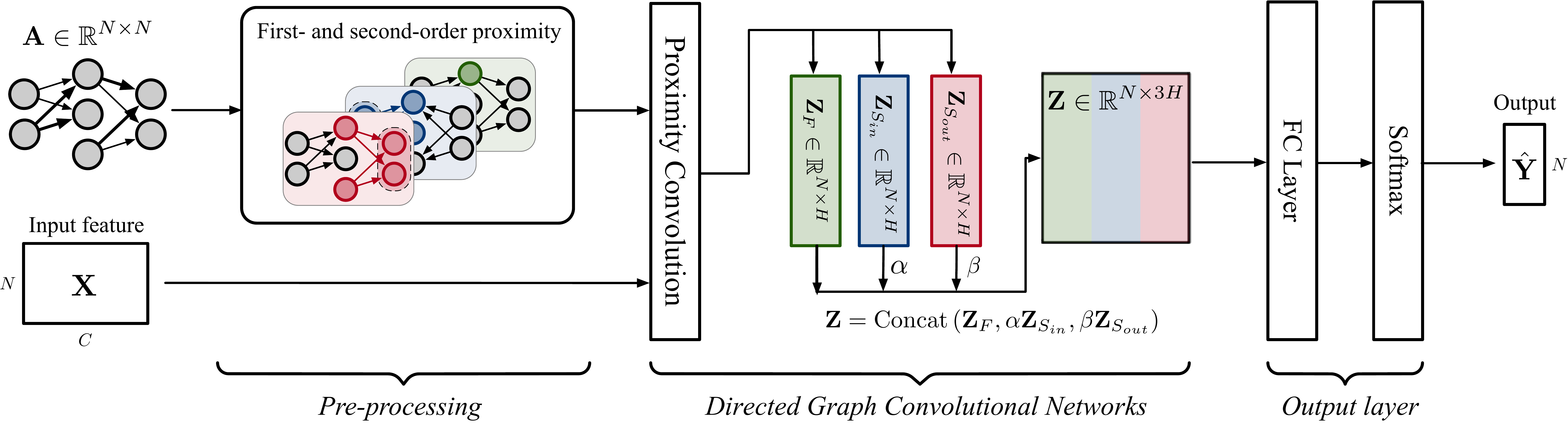}\\
  \setlength{\belowcaptionskip}{-0.2cm} 
  \caption{The schematic depiction of DGCN for semi-supervised learning. Model inputs are an adjacent matrix $\mathbf{A}$ and a features matrix $\mathbf{X}$, while outputs are labels of predict nodes $\mathbf{\hat{Y}}$.}
  \label{model}
\end{figure*}


\subsection{First- and Second-order Proximity}

\label{matrix}

To conduct the feature extraction, we not only obtain the node's features from its directly adjacent nodes, but also extract the hidden information from \emph{second-order} neighbor nodes. Different from other methods considering K-hop neighborhood information\cite{defferrard2016convolutional,abu2018n}, we define new \emph{First- and Second-order Proximity} and show schematically descriptions in Figure \ref{origin_first_in_out}.

\subsubsection{First-order Proximity}
The first-order proximity refers to the local pairwise proximity between the vertices in a graph.


\begin{myDef}
\label{fp}
\textbf{First-order Proximity Matrix}. In order to model the first-order proximity, we define the \emph{first-order} proximity $A_F$ between vertex $v_i$ and $v_j$ for each edge $(i, j)$ in the graph as follows:
\begin{equation}
	A_F (i,j) = A^{sym}(i,j),
\end{equation}
where $\mathbf{A}^{sym}$ is the symmetric matrix of edge weight matrix $\mathbf{A}$. If there is no edge from $v_i$ to $v_j$ or $v_j$ to $v_i$, $A_F(i,j)$ is equal to 0. 

\end{myDef}

In Figure \ref{origin_first_in_out}(b), it is easy to find that vertex $v_1$ has \emph{first-order} proximity with $v_4$. Note that the \emph{first-order} proximity is relaxed for directed graphs. We use the symmetric matrix to replace to original one, which is inevitable losing some directed informations. For this part of the missing information, we will use another way to retain it, which is the \emph{second-order} proximity. This problem does not exist for undirected graphs because its weights matrix is symmetric.

\subsubsection{Second-order Proximity}

The second-order proximity assumes that if two vertices share many common neighbors tend to be similar. In this case, we need to build a second-order proximity matrix, so that similar nodes can be connected with each other.

\begin{myDef}
\label{sp}
\textbf{Second-order Proximity Matrix}. The second-order proximity between vertices is determined by the normalized weights summation of edges linked with their shared neighborhood nodes. In a directed graph $\mathcal{G}$, for vertex $v_i$ and $v_j$, we define the \emph{second-order} in-degree proximity $A_{S_{in}}(i,j)$ and out-degree proximity $A_{S_{out}}(i,j)$:
\begin{equation}
	A_{S_{in}}(i,j) = \sum_{k} \frac{A_{k,i} A_{k,j}}{\sum_{v}A_{k,v}}
\end{equation}

and

\begin{equation}
A_{S_{out}}(i,j) = \sum_{k}\frac{A_{i,k} A_{j,k}}{\sum_{v} A_{v,k}} ,
\end{equation}
where  $\mathbf{A}$ is is the weighted adjacency matrix of $\mathcal{G}$ and $k,v \in \mathcal{V}$. 
\end{myDef}

Since $A_{S_{in}}(i,j)$ sums up the normalized weights of edges which array to both $v_i$ and $v_j$, i.e. $\sum_k A\{i \leftarrow k \rightarrow j \}$, it can best reflect the similarity of the in-degree between vertex $v_i$ and $v_j$. The larger the $A_{S_{in}}(i,j)$, the higher the similarity of the second-order in-degree. Similarly, $A_{S_{out}}(i,j)$ measures the second-order out-degree proximity by accumulating the weights of edges from both $v_i$ and $v_j$, i.e. $\sum_k A\{i \rightarrow k \leftarrow j \}$. If no shared vertices linked from/to $v_i$ and $v_j$, we set their second-order proximity as $0$.

A visualization of these two proximities are shown in Figure \ref{origin_first_in_out}(c). In Figure 2.c.i, $v_1$ and $v_2$ have second-order in-degree proximity, because they share common neighbors $\{1 \leftarrow (5,6) \rightarrow 2 \}$; while $v_1$ and $v_3$ have second-order out-degree proximity, because of $\{1 \rightarrow (7,8) \leftarrow 3 \}$ in the Figure 2.c.ii.

In directed graph, the edges of vertex $v_k$ adjacent to $v_i$ and $v_j$ is pair-wise, thus, $A_{S_{in}}(i,j)=A_{S_{in}}(j,i)$ and $A_{S_{out}}(i,j)=A_{S_{out}}(j,i)$. In other words, $\mathbf{A}_{S_{in}}$ and $\mathbf{A}_{S_{out}}$ are symmetric.


\subsection{Directed Graph Convolution}
\label{dgc}
In the previous section, we define the first-order and second-order proximity, and have obtained three proximity symmetric matrices $\mathbf{A}_F,\mathbf{A}_{S_{in}}$ and $\mathbf{A}_{S_{out}}$. Similar to the authors that define the graph convolution operation on undirected graphs in Section \ref{un}, we use first- and second-order proximity matrix to achieve graph convolution of directed graphs.

In Equation \ref{kipf_complex}, the adjacency matrices $\mathbf{{A}'}$ of the graph stores the information of the graph and provides the receptive field for the filter $\Theta$, so as to realize the transformation from the signal $\mathbf{X}$ to the convolved signal $\mathbf{Z'}$. It is worth noting that the first- and second-order proximity matrix we have defined have similar functions: first-order proximity provides a '1-hop' like receptive field, and second-order proximity provides a '2-hop' like receptive field. Besides, we have the following theorem of first- and second-order proximity:
\begin{mytheo}
\label{theo1}
	The the Laplacian of the first- and second-order proximity $\mathbf{A}_F,\mathbf{A}_{S_{in}}$ and $\mathbf{A}_{S_{out}}$ are positive semi-definite matrices.
\end{mytheo}
\begin{myprof}
	According to Definition \ref{fp} and \ref{sp}, $\mathbf{A}_F,\mathbf{A}_{S_{in}}$ and $\mathbf{A}_{S_{out}}$ are symmetric. We can consider these three matrices as weighted adjacency matrices $\mathbf{A'}$ with non-negative weights of weighted undirected graphs $\mathcal{G'}=(\mathcal{V'},\mathcal{E'})$, so as to formulate Laplacian matrix $\mathbf{L'}$ of $\mathcal{G'}$ in this format:  $\mathbf{L'} = \sum_{uv\in E'} A'_{uv} T_{uv}$, where $e_i \in \mathbb{R}^N$ is the $i^{th}$ standard basis vector, $t_{uv}=e_u-e_v$ and $T_{uv}=t_{uv}{t_{uv}}^T$. Since $T_{uv}$ is positive semi-definite, $\mathbf{L'}$ is a weighted sum of non-negative coefficients with positive semi-definite matrices, which implies $\mathbf{L'}$ is positive semi-definite. $\hfill\blacksquare$ 
\end{myprof}

It should be noted that, the Laplacian of $\mathbf{A'}$ is also positive semi-definite. These features allows us to use these three matrices for spectral convolution. 

\paratitle{Proximity Convolution}

Based on the above analogy, we define the first-order proximity convolution $f_{F}(\mathbf{X},\mathbf{\tilde{A}_F})$, second-order in- and out-degree proximity convolution $f_{S_{in}}(\mathbf{X},\mathbf{\tilde{A}_{S_{in}}})$ and $f_{S_{out}}(\mathbf{X},\mathbf{\tilde{A}_{S_{out}}})$:

\begin{equation}
\label{conv}
\begin{array}{lll}
\mathbf{Z}_F  & = f_{F}(\mathbf{X},\mathbf{\tilde{A}_F})  = & \mathbf{\tilde{D}_F^{-\frac{1}{2}}} ~~\mathbf{\tilde{A}_F} ~~\mathbf{\tilde{D}_F}^{-\frac{1}{2}} \mathbf{X} \Theta\\
\mathbf{Z}_{S_{in}} &= f_{S_{in}}(\mathbf{X},\mathbf{\tilde{A}_{S_{in}}}) = &  \mathbf{\tilde{D}_{S_{in}}}^{-\frac{1}{2}} ~~\mathbf{\tilde{A}_{S_{in}}} ~~\mathbf{\tilde{D}_{S_{in}}}^{-\frac{1}{2}} \mathbf{X} \Theta \\
\mathbf{Z}_{S_{out}} &= f_{S_{out}}(\mathbf{X},\mathbf{\tilde{A}_{S_{out}}})  = &  \mathbf{\tilde{D}_{S_{out}}}^{-\frac{1}{2}} ~~\mathbf{\tilde{A}_{S_{out}}}~~ \mathbf{\tilde{D}_{S_{out}}}^{-\frac{1}{2}} X \Theta \\
\end{array},
\end{equation}
where adjacent weighted matrix $\mathbf{\tilde{A}}$ ($\mathbf{A}$ added self-loop) is used in the definition to derive $\mathbf{\tilde{A}_F}$, $\mathbf{\tilde{A}_{S_{in}}}$ and $\mathbf{\tilde{A}_{S_{out}}}$. $\mathbf{\tilde{D}_{F}}=diag(\sum_{j}^{n}{\tilde{A}_F}(i,j))$, $\mathbf{\tilde{D}_{S_{in}}}=diag(\sum_{j}^{n}{\tilde{A}_{S_{in}}}(i,j))$ and $\mathbf{\tilde{D}_{S_{out}}}=diag(\sum_{j}^{n}{\tilde{A}_{S_{out}}}(i,j))$.


It can be seen that $\mathbf{Z}_F, \mathbf{Z}_{S_{in}}$ and $\mathbf{Z}_{S_{out}}$ not only obtain rich first- and second-order neighbor feature information, but  $\mathbf{Z}_{S_{in}} $ and $ \mathbf{Z}_{S_{out}}$ also retain the directed graph structure information. Based on these facts, we further design a fusion method to integrate the three signals together, so as to retain the characteristics of the directed structure while obtaining the surrounding information.



\paratitle{Fusion Operation}

Directed graph fusion operation $\Gamma$ is a signal fusion function of the first-order proximity convolution output $\mathbf{Z}_F$, second-order in- and out-degree proximity convolution outputs $\mathbf{Z}_{S_{in}}$ and $\mathbf{Z}_{S_{out}}$:

\begin{equation}
	\mathbf{Z} = \Gamma (\mathbf{Z}_F,\mathbf{Z}_{S_{in}},\mathbf{Z}_{S_{out}}).
\end{equation}

Fusion function $\Gamma$ can be various, such as normalization functions, summation functions, and concatenation. In practice, we find concatenation fusion has the best performance. A simple example is:

\begin{equation}
	\mathbf{Z} = Concat( \mathbf{Z}_F , \alpha \mathbf{Z}_{S_{in}} , \beta  \mathbf{Z}_{S_{out}}),
\end{equation}
where $Concat(\cdot)$ is the concatenation of matrices, $\alpha$ and $\beta$ are weights to control the importance between different proximities. For example, in a graph with fewer second-order neighbors, we can reduce the values of $\alpha$ and $\beta$ and use more first-order information. $\alpha$ and $\beta$ can be set manually or trained as learnable parameters.

For a given features matrix $\mathbf{X}$ and a directed adjacency matrix $\mathbf{A}$, after taking all the steps above, we can get the final directed graph result $Z = f(\mathbf{X},A)$.

\subsection{Implementation}
\label{implement}

In the previous section, we proposed a simple and flexible model on the directed graph, which can extract the surrounding information efficiently and retain the directed structure. In this section, we will implement our model to solve semi-supervised node classification task. More specifically, how to mine the similarity between node class using weighted adjacency matrix $\mathbf{A}$ when there is no graph structure information in node feature matrix $\mathbf{X}$. 

For this task, we build a two layer network model on directed graph with a weighted adjacency matrix $\mathbf{A}$ and node feature matrix $\mathbf{X}$, is schematically depicted in Figure \ref{model}. In the first step, we calculate the first- and second-order proximity matrixes $\mathbf{\hat{A}_F}$,$\mathbf{\hat{A}_{S_{in}}}$ and $\mathbf{\hat{A}_{S_{out}}}$ according to Equation \ref{eq1} in the preprocessing stage. Our model can be written in the following form of forward propagation: 

\begin{equation}
\begin{array}{l}

\mathbf{\hat{Y}} =f(\mathbf{X},\mathbf{A})=softmax \begin{pmatrix} Concat\begin{pmatrix} ReLU \begin{pmatrix} 
~~~\mathbf{\hat{A}_F} ~~~\mathbf{X} ~~~\Theta^{(0)}\\
\alpha ~~~~~\mathbf{\hat{A}_{S_{in}}} ~~~\mathbf{X}  ~~~\Theta^{(0)}\\
\beta ~~~~~\mathbf{\hat{A}_{S_{out}}} ~~~\mathbf{X} ~~~\Theta^{(0)}

\end{pmatrix}
\Theta^{(1)} 
\end{pmatrix}
\end{pmatrix}
\end{array}.
\end{equation}
In this formula, the first layer is the directed graph convolution layer. Three different proximity convolutions share a same filter weight matrix $\Theta^{(0)} \in \mathbb{R}^{C \times H}$, which can transform the input dimension $C$ to the embedding size $H$. After feature matrix $\mathbf{X}$ through the first layer, there will be three different convolved results. Then we use a fusion function to concatenate them together, $\alpha$ and $\beta$ are variable weights to trade off first- and second-order feature embedding. The second layer is a fully connected layer, which we use to change feature dimension from $3H$ to $F$. $\Theta^{(1)}\in \mathbb{R}^{3H \times F} $ is an embedding-to-output weight matrix. The softmax function is defined as $\operatorname{softmax}\left(x_{i}\right)=\frac{1}{\mathcal{Z}} \exp \left(x_{i}\right)$ with $\mathcal{Z} = \sum_{i} \exp \left(x_{i}\right)$. We use all labeled examples to evaluate the cross-entropy error for semi-supervised node classification task:

\begin{equation}
	\mathcal{L}=-\sum_{l \in \mathbf{V}_{L}} \sum_{f=1}^{F} Y_{l f} \ln Z_{l f}
\end{equation}
where $Y_i$ is the label of vertex $i$ and $\mathcal{V}_{L}$ is the subset of $\mathcal{V}$ which is labeled. The pseudocode of DGCN is attached in the Appendix.

\subsection{Discussion}
\label{diss}

\subsubsection{Time and Space Complexity}

For graph convolution defined in the Equation \ref{conv}, we can use a sparse matrix to store weighted adjacency matrix $\mathbf{A}$. Because we use full batch training in this task, full dataset has to be loaded into memory for every iteration. The memory space cost is $\mathcal{O}(|\mathcal{E}|)$, which means it is linear with the number of edges. 

At the same time, we use the sparse matrix and the density matrix to multiply during the convolution operation. The multiplication of the sparse matrix can be considered to be linearly related to the number of edges $|\mathcal{E}|$. In the semi-supervised classification task, we need to multiply with $\Theta^{0}  \in \mathbb{R}^{C \times H}$ and $\Theta^{1} \in \mathbb{R}^{3H \times F} $. Thus, we can obtain the computational complexity of the model as $\mathcal{O}(|\mathcal{E}|CHF)$.

\subsubsection{Generalization to other Graph Models}
\label{gene}
Our method using first-and second-order proximity to improve the convolution receptive field and retain directed information has strong generalization ability. In most spectral-based models, we can use these proximity matrices to replace the original adjacency matrix.

Take Simplifying Graph Convolutional Networks (SGC)\cite{wu2019simplifying} as an example, we can generalize our method to the SGC model as follows:
\begin{equation}
	\hat{\mathbf{Y}}_{\mathrm{S'}}=\operatorname{softmax}\left(\mathbf{S'} \mathbf{X} \Theta\right),
\end{equation}
where we use the concatenation of first- and second-order proximity matrix $\mathbf{A_{F}}$,$\mathbf{A_{S_{in}}}$ and $\mathbf{A_{S_{out}}}$ to replace the origin $K$-th power of adjacency matrix, $\mathbf{S}^{K}$, $\mathbf{S}$ is the simplified adjacency matrix defined in SGC\cite{wu2019simplifying}. Experimental results in Section \ref{sgc} show that integrating our method can not only make SGC model simpler, but also improve accuracy.
\subsubsection{Relation with K-hop Methods}

Our work considers not only the first-order relationships, but also the second-order ones when extracting surrounding information. The first-order proximity has the similar function to the 1-hop, which is to obtain the information of directly connected points. However, the reason why we do not define our second-order relationship as 2-hop is that it does not need node $i$ and node $j$ to have a 2 degree path directly. 

For the K-hop method, and $K=2$, they need a $K$ degree path from $i$ to $j$, i.e. $\{i \rightarrow k \rightarrow j \}$ in directed graph. However, in our method, the second-order pattern diagram is transformed into $\{i \rightarrow k \leftarrow j \}$ and $\{i \leftarrow k \rightarrow j \}$, which is obvious that we get information from the shared attributes among nodes, not from the path. What's more, when evaluating the second-order proximity of nodes, we do not use the weights of the connecting edges between them, but use the sum of the normalized weights of their shared nodes.

\section{Experiments}

In this section, we evaluate the effectiveness of our model using experiments. We test on citation and co-purchase networks, and then evaluate the performance on directed and undirected dataset.

\subsection{Datasets and Baselines}

We use the several datasets to evaluate our model. In the citation network datasets: \textsc{Cora-Full} \cite{bojchevski2017deep}, \textsc{Cora-ML} \cite{bojchevski2017deep}, \textsc{CiteSeer} \cite{sen2008collective} , \textsc{DBLP} \cite{pan2016tri} and \textsc{PubMed} \cite{namata2012query}, nodes represent articles, while edges represent citation between articles. These datasets also include bag-of-words feature vectors for each article. In addition to the citation network, we also use the Amazon Co-purchase Network: \textsc{Amazon-Photo} and \textsc{Amazon-Computers} \cite{shchur2018pitfalls}, where nodes represent goods, while edges represent two kinds of goods that are often purchased together. Bag-of-words encoded product reviews product category are also given as features, and class labels are given by the product category. In the above datasets, except \textsc{DBLP} and \textsc{PubMed} are undirected data we obtained, the rest are directed. The statistics of datasets are summarized in Appendix.

We compare our model to five state-of-the-art models that can be divided into two main categories: 1) \textbf{spectral-based} GNNs including ChebNet \cite{defferrard2016convolutional}, Graph Convolutional Networks (GCNs) \cite{kipf2016semi},  Simplifying Graph Convolutional Networks (SGC) \cite{wu2019simplifying} and 2) \textbf{spatial-based} GNNs containing GraphSage \cite{hamilton2017inductive} and Graph Attention Networks (GAT) \cite{velivckovic2017graph}. The descriptions and settings of them are introduced in the Appendix.


\subsection{Experimental Setup}

\begin{table*}[t]
\caption{Mean test accuracy and standard deviation in percent. Underlined bold font indicates best results.\label{result}}
\small
\begin{tabular}{|c||c||c||c||c||c||c||c||c||c|}
\hline
\multirow{2}{*}{\textbf{\emph{Label Split}}} & \multirow{2}{*}{$\quad$\textsc{Cora-Full}$\quad$} & \multirow{2}{*}{$\quad$\textsc{Cora-ML}$\quad$} & \multirow{2}{*}{$\quad$\textsc{CiteSeer}$\quad$} & \multirow{2}{*}{$\qquad$\textsc{DBLP}$\qquad$}  & \multirow{2}{*}{$\qquad$\textsc{PubMed}$\qquad$} & \multirow{2}{*}{\textsc{Amazon-Photo}} & \multirow{2}{*}{\textsc{Amazon-Computer}} \\
 &  &  &  &  &  &  &  \\ \hline
 \hline
 
ChebNet  & $58.0 \pm 0.5 $ &$ 79.2 \pm 1.4$ & $59.7 \pm 4.0$ & $64.0 \pm 2.8$ & $74.6 \pm 2.5$ & $82.5 \pm 2.4$ & $72.9 \pm 3.0$ \\ \hline
GCN      & $59.1 \pm 0.7$ & $81.7 \pm 1.2$ & $64.7 \pm 2.3$& $71.5  \pm 2.7$& $76.8 \pm 2.2 $& $90.4 \pm 1.5 $&  $81.9 \pm 1.9$ \\ \hline
SGC       & \underline{\textbf{61.2 $\pm$ 0.6 }}& $80.3 \pm 1.1$& $61.4 \pm 3.4$& $69.2 \pm 2.8$& $75.8 \pm  2.8$& $89.4 \pm  1.4$& $80.2 \pm 1.2$\\ \hline
GraphSage & $58.1 \pm 0.7$ & $80.2 \pm 1.6$& $62.8 \pm 2.1$& $68.1 \pm 2.5$& $75.2 \pm 3.2$& $89.8 \pm 1.9$& $80.4 \pm 2.5$\\ \hline
GAT       & $60.8 \pm 0.6$& $81.5 \pm 1.0$& $63.7 \pm 2.0$& $71.8  \pm 2.6$& $76.5 \pm 2.3$& $90.0 \pm 1.3$ & $81.2 \pm 2.5$\\ 

\hline
\textbf{DGCN} & $60.8 \pm 0.6$ & \underline{\textbf{ 82.0 $\pm$ 1.4 }}& \underline{\textbf{65.4 $\pm$ 2.3}} & \underline{\textbf{72.5 $\pm$ 2.5}} & \underline{\textbf{76.9 $\pm$ 1.9}} & \underline{\textbf{90.8 $\pm$ 1.1}} & \underline{\textbf{82.0 $\pm$ 1.7}}\\ \hline
\end{tabular}

\setlength{\belowcaptionskip}{-0.5cm} 
\end{table*}

\begin{figure*}[t]
\centering
\subfigure[GCN]{
\begin{minipage}[t]{0.3\linewidth}
\centering
\includegraphics[width=4.2cm]{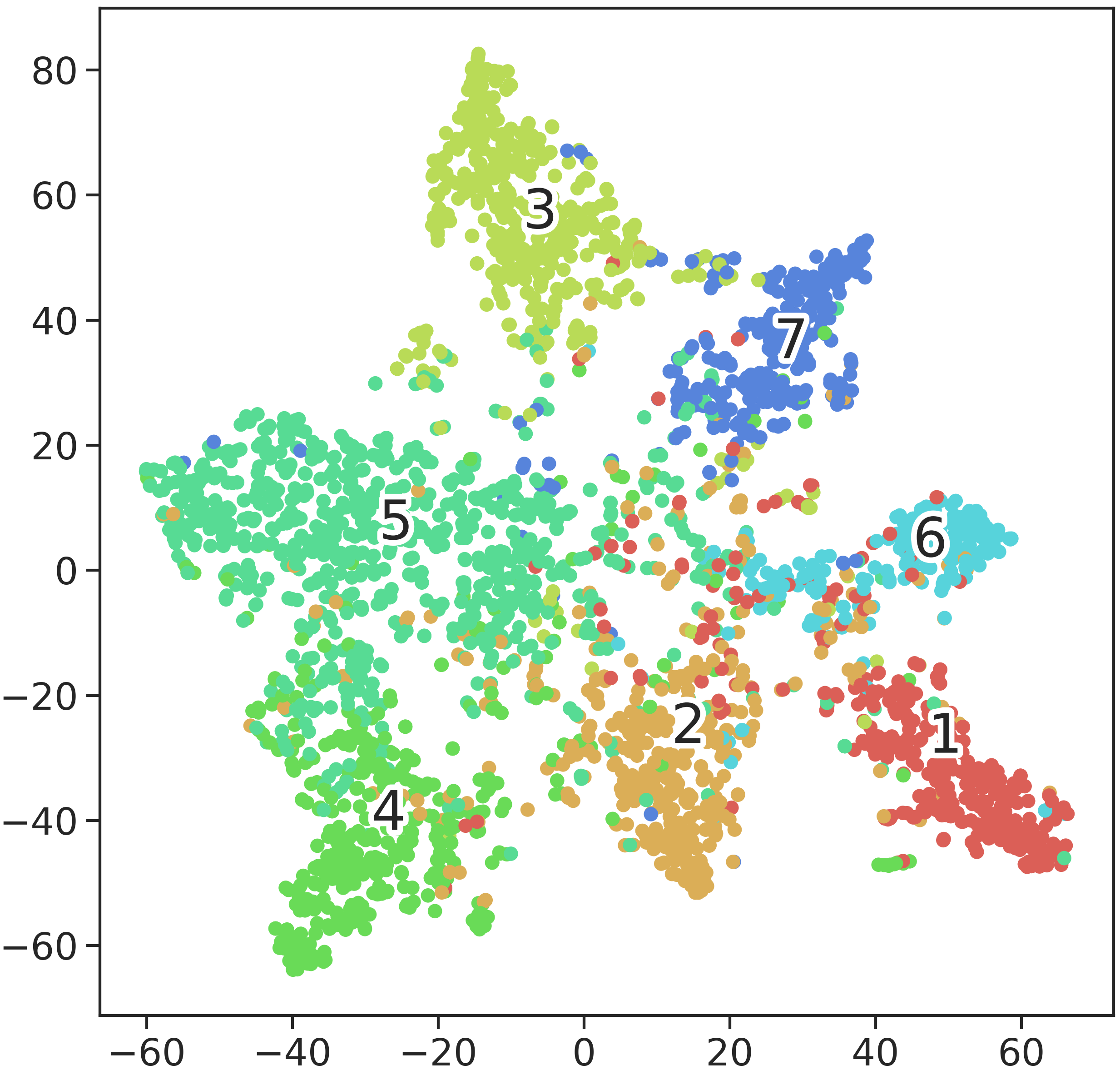} 
\end{minipage}
}
\subfigure[DGCN uses only First-order Proximity]{
\begin{minipage}[t]{0.3\linewidth}
\centering
\includegraphics[width=4.25cm]{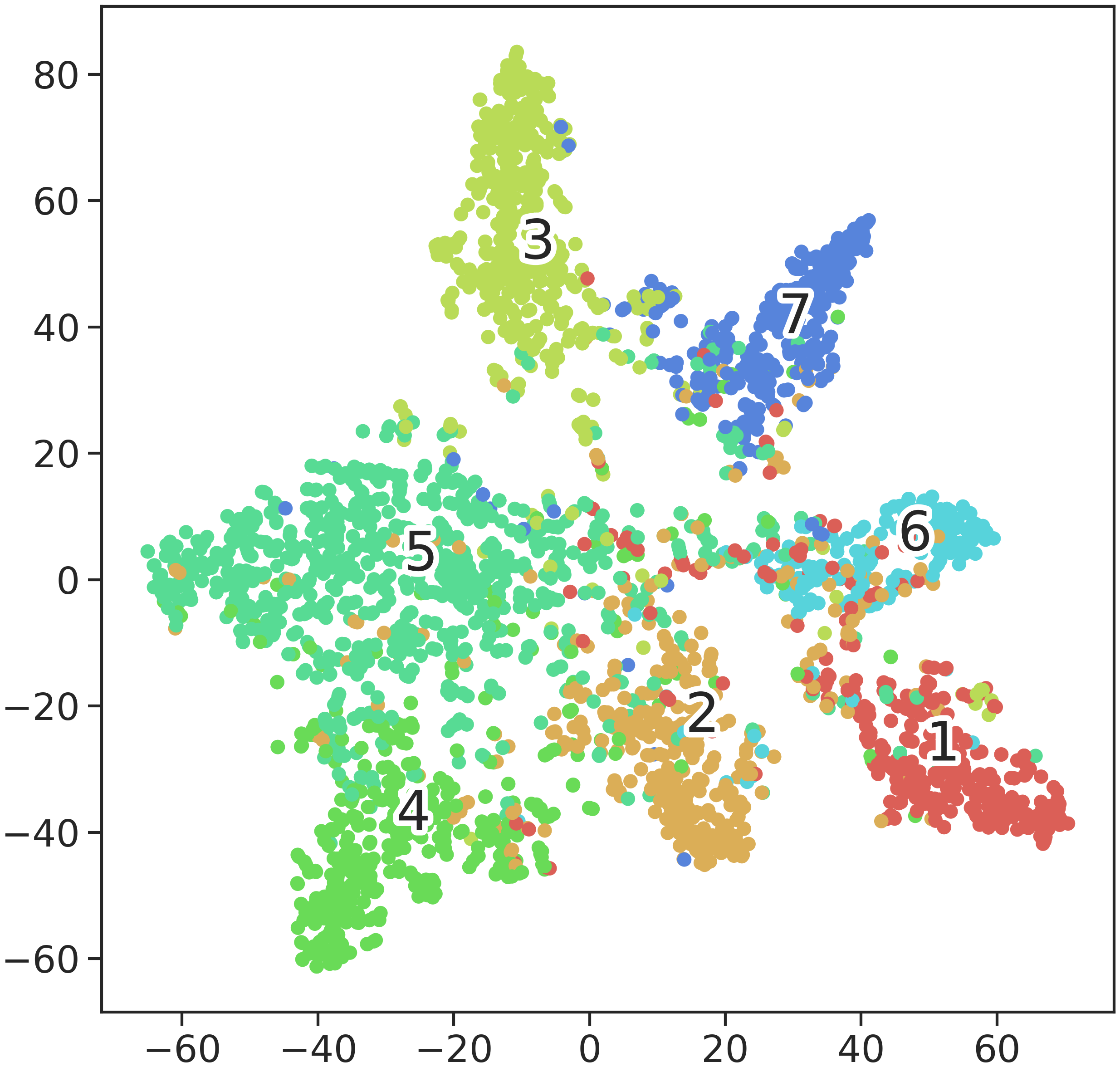} 
\end{minipage}
}
\subfigure[DGCN uses both First- and Second-order Proximity]{
\begin{minipage}[t]{0.3\linewidth}
\centering
\includegraphics[width=4.2cm]{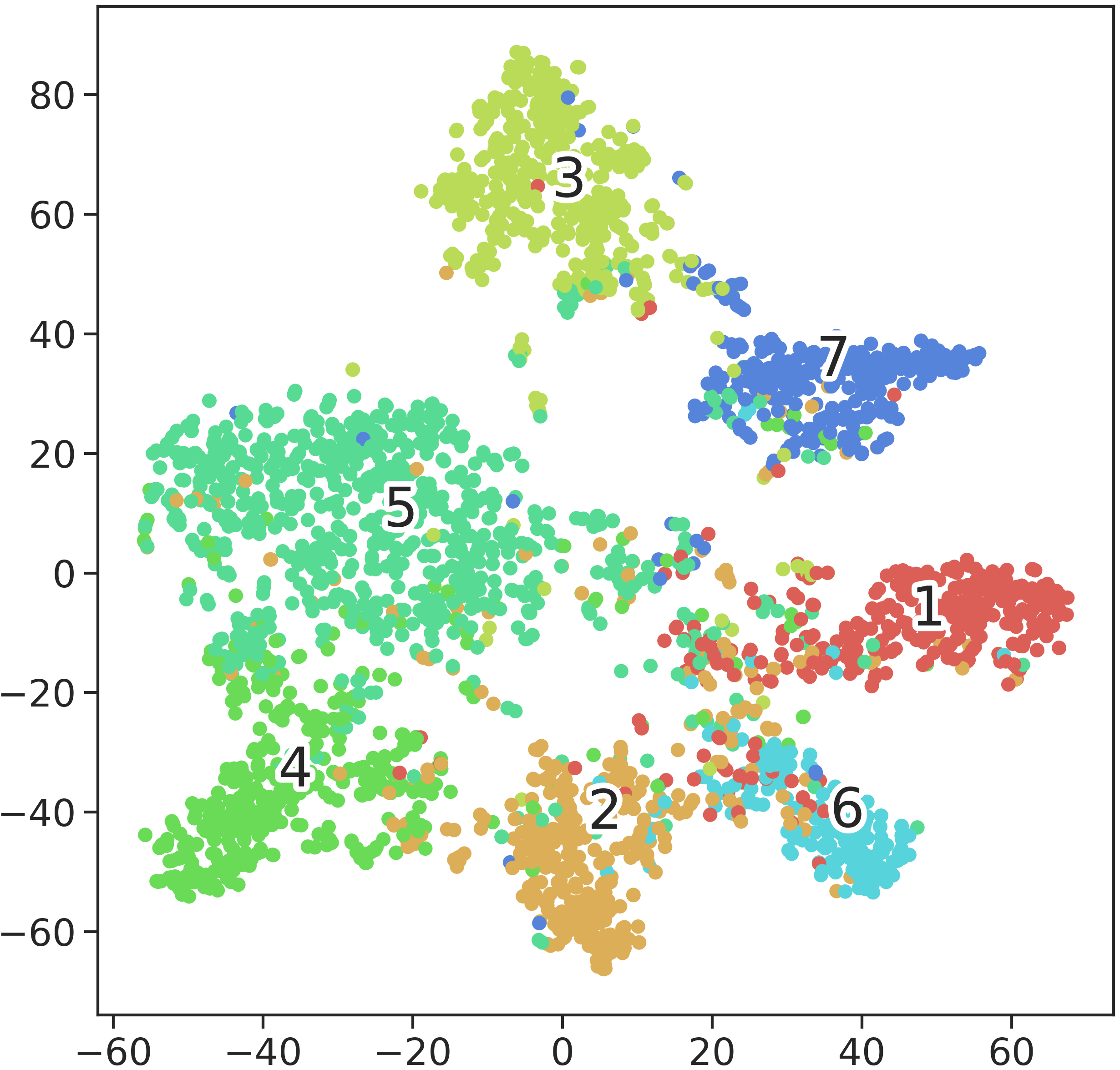} 
\end{minipage}
}
\caption{2D t-SNE\cite{maaten2008visualizing} visualizations of the first convolutional layer feature outputs on \textsc{Cora-ML} dataset. The data of different classes (denote by colors) are distributed more clearly and compactly in our model feature map \textbf{(c)}.\label{tsne}}
\end{figure*}

\paratitle{Our Method Setup} We train the two-layer DGCN model built in Section \ref{implement} for semi-supervised node classification task and provide additional experiments to explore the effect of model layers on the accuracy. We use full batch training, and each iteration will use the whole dataset. For each epoch, we initialize weights according to Glorot and Bengio\cite{glorot2010understanding} and initialize biases with zeros. We use Adam\cite{kingma2014adam} as optimizer with a learning rate of 0.01. Validation set is using for hyperparameter optimization, which have weights( $\alpha, \beta$) of first- and second-order proximity, dropout rate for all layer, L2 regularization factor for the DGCN layer and embedding size.

\paratitle{Dataset Split} The split of the dataset will greatly affect the performance of the model\cite{shchur2018pitfalls,klicpera2018predict}. Especially for a single split, not only will it cause overfitting problems during training, but it is also easy to get misleading results. In our experiments,  we will randomly split the data set and perform multiple experiments to obtain stable and reliable results. What's more, we also test the model under different sizes of training set in Section \ref{results}. For train/validation/test splitting, we choose 20 labels per class for training set, 500 labels for validation set and rest for test set, which follows the split in GCN\cite{kipf2016semi}, which marked as \textbf{\emph{Label Split}}. 



\subsection{Experimental Results}
\label{results}

\paratitle{Semi-Supervised Node Classification}

The comparison results of our model and baselines on seven datasets are reported in Table~\ref{result}. Reported numbers denote classification accuracy in percent. Except \textsc{DBLP} and \textsc{PubMed}, all other datasets are directed. We train all models for a maximum of 500 epochs and early stop if the validation accuracy does not increase for 50 consecutive epochs in each dataset split, then calculate mean test accuracy and standard deviation averaged over 10 random train/validation/test splits with 5 random weight initializations. We use the following settings of hyerparameters for all datasets: drop rate is 0.5; L2 regularization is $5\cdot10^{-4}$; $\alpha = \beta =1$ (we will explain the reasons in later section). Besides, we choose embedding size as 128 for Co-purchase Network: \textsc{Amazon-Photo} and \textsc{Amazon-Computer}, and 64 for others.

Our method achieved the state-of-the-art results on all datasets except \textsc{Cora-Full}. Although SGC achieves the best results on \textsc{Cora-Full}, its performances on other datasets are not outstanding. Our method achieves best results on both directed (\textsc{Cora-ML}, \textsc{CiteSeer}) and undirected (\textsc{DBLP}, \textsc{PubMed}) datasets. Our method is not significantly improved compared to GCN on the \textsc{Amazon-Photo} and \textsc{Amazon-Computer}, mainly because our model has only one convolutional layer while GCN uses two convolutional layers. The single layer network representation capability is not enough to handle large nodes graph.

\paratitle{First- \& Second-order Proximity Evaluation}

Table \ref{smooth} reports the two smoothness values of \textsc{Cora-ML}, \textsc{CiteSeer}, \textsc{DBLP} and \textsc{PubMed}. \emph{Features Smoothness} and \emph{Label Smoothness} are defined in Preliminaries \ref{for} to measure the quantity and quality of information that models can obtain, respectively. After adding second-order proximity, the feature smoothness of \textsc{CiteSeer} increases from $8.719 \times 10^{-4}$ to $54.720 \times 10^{-4}$, while the label smoothness increases from $0.4893$ to $0.5735$. This change shows that the second-order proximity is very effective on this dataset, which help increase the quantity and improve quality of information from the surrounding. The label smoothness of other datasets decrease slightly, while their feature smoothness significantly increase. In other words, the second-order proximity widens the receptive field, thus greatly increases the amount of information obtained. Besides, the second-order proximity preserves the directed structure information which helps it to filter out valid information. The t-SNE results shown in Figure \ref{tsne} also prove that second-order proximity can help the model achieve better classification results.
\begin{table}[h]
\caption{Smoothness values for First- and Second-order Proximity on different datasets. ${1^{st}}$ represents first-order proximity, $1^{st}\&2^{nd}$ represents first- and second-order proximity, $\lambda_f$ means Feature Smoothness and $\lambda_l$ means Label Smoothness.}
\small
\label{smooth}
\begin{tabular}{|c||c||c||c||c||c||c|}
\hline
\multirow{2}{*}{\textbf{\emph{Smoothness}}} & \multirow{2}{*}{\textsc{Cora-ML}} & \multirow{2}{*}{\textsc{CiteSeer}} & \multirow{2}{*}{\textsc{DBLP}}  & \multirow{2}{*}{\textsc{PubMed}}\\
 &  &  &  & \\ \hline
 \hline

$1^{st}~~~\lambda_f(10^{-4})$& 3.759  & 8.719 & 3.579 & 3.135 \\ \hline
${1^{st}\& 2^{nd}} ~~\lambda_f(10^{-4})$& 9.789 & 54.720 & 36.810 & 20.480 \\ \hline \hline
$1^{st}~~~\lambda_l$& 0.577     & 0.489     & 0.656     & 0.605     \\ \hline 
${1^{st}\& 2^{nd}} ~~\lambda_l$& 0.393     & 0.574    & 0.459     & 0.547  \\ \hline

\end{tabular}
\end{table}

\paratitle{Generalization to other Model (SGC)}
\label{sgc}

In order to test the generalization ability of our model, we design an experiment according to the scheme proposed in Section \ref{gene}. We use the concatenation of first- and second-order proximity matrix to replace the origin $K$-th power of adjacency matrix. The generalized SGC model is denoted by \textbf{SGC+DGCN}. In addition, we set the power time $K=2$ to the origin SGC model. We follow the experimental setup described in the previous section, and the results are summarized in Table \ref{gsgc}. Obviously, our generalized model outperforms the original model on all datasets, not only significantly improves classification accuracy, but also has more stable performance (with smaller standard deviations). Our method has good generalization ability because it has a simple structure that can be plugged into existing models easily while providing a wider receptive field by the second-order proximity matrix to improve model performance.

\begin{table}[h]
\caption{Accuracy of origin SGC and generalized SGC. Underlined bold font indicates best results.}
\label{gsgc}
\small
\begin{tabular}{|c||c||c||c||c||c||c|}
\hline
\multirow{2}{*}{\textbf{\emph{Label Split}}}  & \multirow{2}{*}{\textsc{Cora-ML}} & \multirow{2}{*}{\textsc{CiteSeer}} & \multirow{2}{*}{\textsc{DBLP}}  & \multirow{2}{*}{\textsc{PubMed}}\\
   &  &  &  & \\ \hline
 \hline
 
SGC       & $80.3 \pm 1.1$& $61.4 \pm 3.4$& $69.2 \pm 2.8$& $75.8 \pm  2.8$\\ \hline

\textbf{SGC+DGCN} & \underline{\textbf{82.3 $\pm$ 1.4}} & \underline{\textbf{63.8 $\pm$ 2.0}} & \underline{\textbf{71.1 $\pm$ 2.3}} & \underline{\textbf{76.5 $\pm$ 2.3}} \\ \hline
\end{tabular}
\setlength{\belowcaptionskip}{-0.2cm} 
\end{table}

\paratitle{Effects of Model Depth}

\begin{figure}[!t]
  \centering
  \includegraphics[width=8cm]{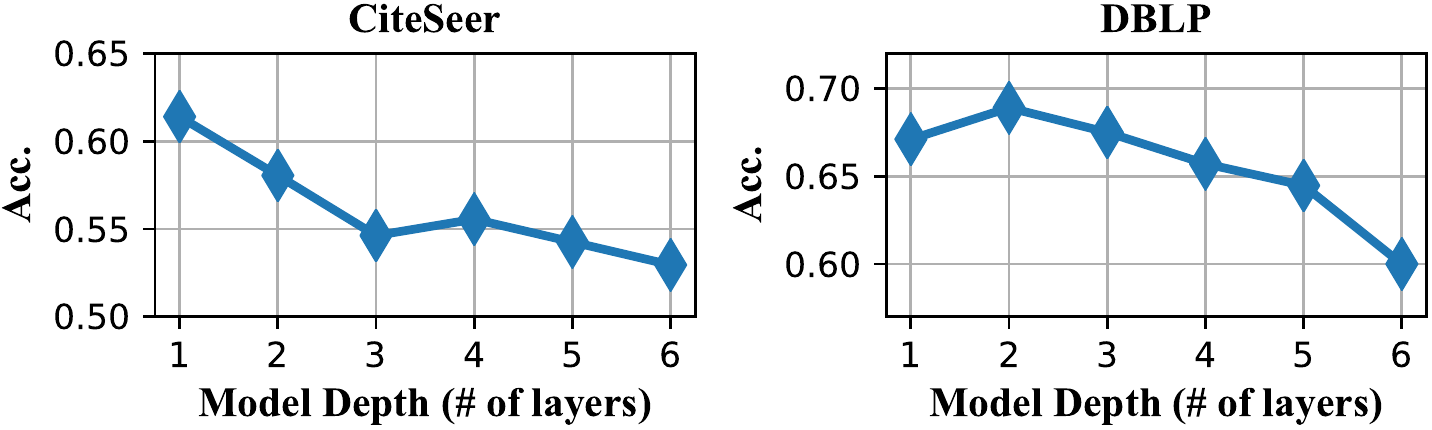}\\
  \setlength{\belowcaptionskip}{-0.5cm} 
  \caption{Effects to classification accuracy when DGCN goes deeper on  \textsc{CiteSeer} and \textsc{DBLP}.}
  \label{depth}
\end{figure}

We investigate the effects of model depth (number of convolutional layers) on classification performance. To prevent overfitting with only one layer, we increase the difficulty of the task and set the training set size per class to 10, validation set size to 500 and the rest as test set. The other settings are the same with previous. Results are summarized in Figure \ref{depth}. Obviously, for the datasets experimented here, the best results are obtained by a 1- or 2-layer model and test accuracy does not increase when the model goes deeper. The main reason is overfitting. The increase in the number of model layers not only greatly increases the amount of parameters, but also widens the receptive field of the convolution operation. When DGCN has only one layer, we only need to obtain information from surrounding connected nodes and nodes of shared 1-hop neighbors. When the DGCN changes to $K$ layers, we need consider both K-hop neighborhoods and the points that share the K-hop neighbors. For a simple semi-supervised learning task, deep DGCN obtains too much information, which easily leads to overfitting.

\paratitle{Weights Selection of First- and Second-order Proximity}

\begin{figure}[h]
  \centering
  \includegraphics[width=8cm]{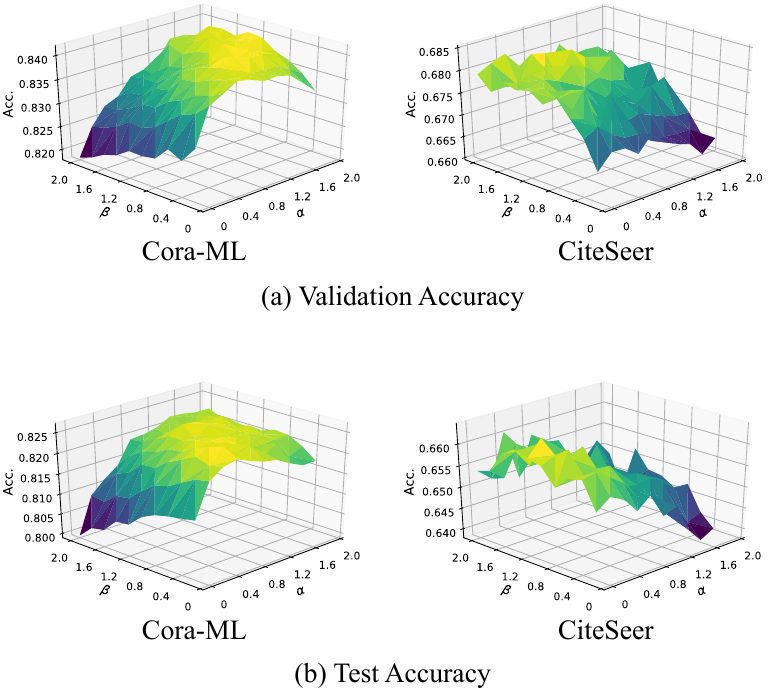}\\
  \setlength{\belowcaptionskip}{-0.5cm} 
  \caption{Accuracy in validation and test set for different weights ($\alpha$ and $\beta$) selection on \textsc{Cora-ML} and \textsc{CiteSeer}.}
  \label{diff_ab}
\end{figure}

We set two hyperparameters $\alpha$ and $\beta$ defined in Section \ref{dgc} to adjust the first- and second-order proximity weights when concatenating them. Figure \ref{diff_ab} shows the accuracy in validation and test set with different weights. We set $\alpha$ and $\beta$ change within $(0,2]$. We find that when the hyperparameters take the boundary values, the accuracies decreases significantly. When the values of the two hyperparameters are close, the accuracies of the model will increase. This is because the second-order in-degree and out-degree proximity matrix not only represent the relationship between the nodes' shared neighbors, but also encode the structure information of the graph, which needs both in- and out-degree matrix. Besides, the unbalance of second-order in- and out-degree makes the optimal hyperparameters combination differ for datasets. Therefore, we use a combination $\alpha = \beta =1$ that can achieve balanced performance.

\paratitle{Effects of Training Set Size}

\begin{figure}[t]
  \centering
  \includegraphics[width=8cm]{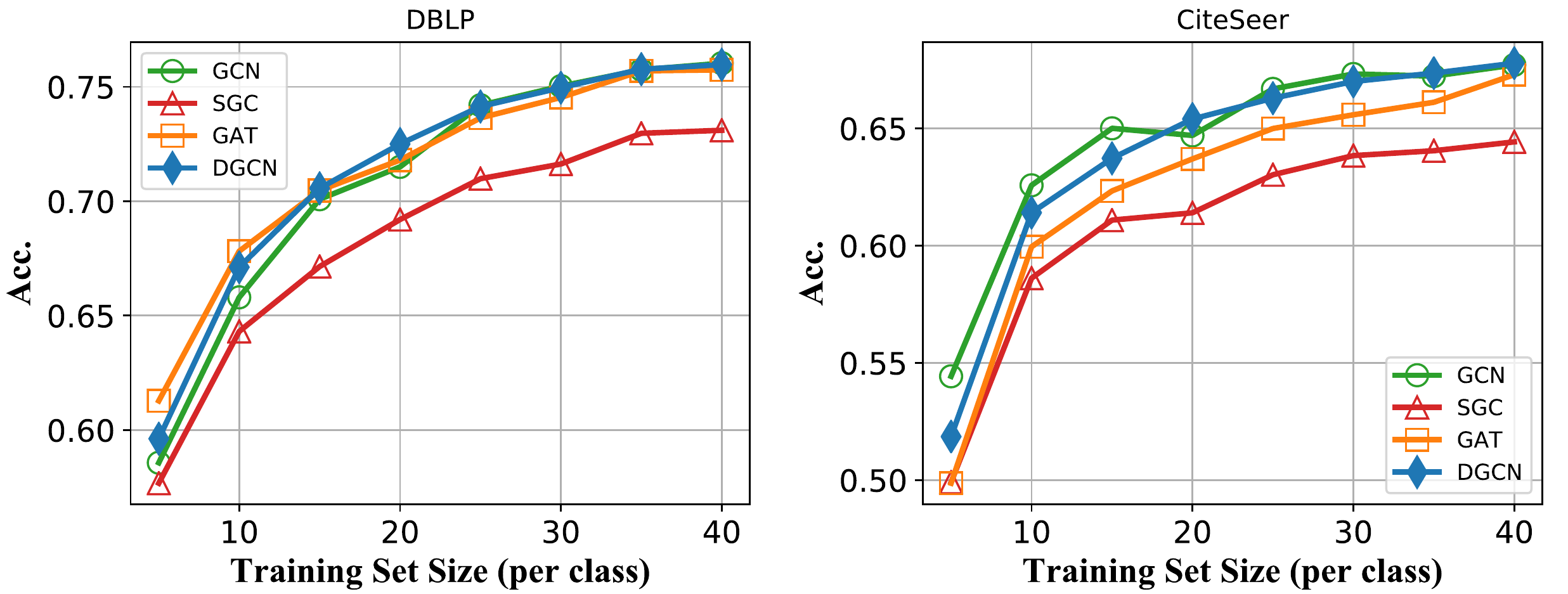}\\
  \setlength{\belowcaptionskip}{-0.5cm} 
  \caption{Accuracy for different training set sizes (number of labeled nodes per class) on \textsc{DBLP} and \textsc{CiteSeer}.}
  \label{size}
\end{figure}

Since the label rates for real world datasets are often small, it is important to study the performance of the model on small training set. Figure \ref{size} shows how the number of training nodes per class to affect accuracy of different models on \textsc{Cora-ML} and \textsc{DBLP}. These four methods perform similarly under small training set size. As the amount of data increases, the accuracy greatly improves. Our method does not perform as well as GCN on \textsc{CiteSeer} and GAT on \textsc{DBLP} respectively. This can be attributed to the second-order proximity and model structure. In the case of less training data, the second-order proximity matrix will become very sparse, which makes it unable to supplement sufficient information. And our model has only one layer of convolution structure, which is not effective when we can not get enough information. On the country, GAT uses fixed eight attention heads and GCN uses two convolutional layers to aggregate node features.

\section{Related Work}

The field of representation learning of graph data is evolving quickly, a lot of works have contributed to it in recent years. We focus on the models similar to our method.


\paratitle{First- and Second-order Proximity Related}
Previous works have already found the powerful information extraction capabilities of second-order proximity. Zhou et al. \cite{zhou2005semi} propose a regularization framework for semi-supervised classification problem, which represents directed graph in the form of bipartite graph and design smoothness function based on the $hub$ and $authority$ model. Tang et al.\cite{tang2015line} propose an efficient graph embedding model for large-scale information network using first- and second-order proximity to retain graph structure information. Besides, SDNE\cite{wang2016structural} designs a semi-supervised deep model for network embedding, which combines first- and second-order proximity components to preserves highly non-linear network structure. Different from our model, when they define the first- and second-order proximity between nodes, they also consider the similarity between node feature vectors, which is not applied in our model.

\paratitle{K-hop Method Related}
Another common way to get more node information is K-hop described in many previous articles. It's used to improve the convolution receptive field. For ChebyNets\cite{defferrard2016convolutional}, when set the $K=2$, the convolution kernel can extract the information of the second-degree node from the central node. Another work is N-GCN\cite{abu2018n}, the researchers inspire from random walk, they build a multi-scale GCN which uses different powers of adjacency matrices $\mathbf{A}$ as input to achieve extract feature from different k-hop neighborhoods. It can gain the information from the $k^{th}$ step from current node, which the same idea with K-hop. However, their K-hop methods are only applicable to undirected graphs, and our method is applicable to both types of graphs.

\section{Conclusion and Future Work}

In this paper, we present a novel graph convolutional networks \textbf{DGCN}, which can be applied to the directed graphs. We define \emph{first- and second-order} proximity on the directed graph to enable the spectral-based GCNs to generalize to directed graphs. It can retain directed features of graph and expand the convolution operation receptive field to extract and leverage surrounding information. Besides, we empirically show that this method helps increase the quantity and improve quality of information obtained. Finally, we use semi-supervised classification tasks and extensive experiments on various real-world datasets to validate the effectiveness and generalization capability of \emph{first- and second-order} proximity and the improvements obtained by DGCNs over other models.

Currently, our approach is able to effectively learn directed graph representation by fused first- and second-order proximity matrices. While the selection of fusion function and concatenation weights are still manually. In the future, we will consider how to design a more principled way for fusing proximities matrices together automatically. We will also consider adapting our approach to mini-batch training in order to speed up large dataset training. In addition, we will study how to generalize our model to inductive learning.

%

\bibliographystyle{ACM-Reference-Format}
\bibliography{references}
\appendix

%
%
%
%

\newpage

%
%
%
%

\section{REPRODUCIBILITY DETAILS}
To support the reproducibility of the results in this paper, we details the pseudocode, the datasets and the baseline settings used in experiments. We implement the DGCN and all the baseline models using the python library of PyTorch \footnote{https://pytorch.org} and DGL 0.3 \footnote{https://www.dgl.ai}. All the experiments are conducted on a server with one GPU (NVIDIA GTX-2080Ti), two CPUs (Intel Xeon E5-2690 * 2) and Ubuntu 16.04 System.

\subsection{DGCN pseudocode}
%
%

\begin{algorithm}  

\caption{DGCN procedure}  
\LinesNumbered  
\KwIn{Graph: ${\mathcal{G}} = (\mathcal{V}, \mathcal{E})$; \newline
graph adjacency matrix: $\mathbf{A}$;\newline
    features matrix: $\mathbf{X}$; \newline 
    non-linear function: $\sigma$; \newline
    weight matrices: $\mathbf{\Theta}$;\newline
    concat weight: $\alpha$, $\beta$}
\KwOut{Predict class matrix $\mathbf{\hat{Y}}$}  
\textbf{Initialize} $\mathbf{\Theta}$ ;

\underline{First- and Second-order Proximity Computation}\newline
\For{$i \in \mathcal{V}$}{
	\For{$j \in \mathcal{V}$}{
	$A_F(i,j)=A^{sym}(i,j)$ \newline
	${SUM}_{in} \gets 0$ \newline
	${SUM}_{out} \gets 0$ \newline
	\For{$k \in \mathcal{V}$}{
	\If {$(k,i) \& (k,j) \in \mathcal{E}$ }{
             ${SUM}_{in} \gets {SUM}_{in}+ \frac{A_{k,i} A_{k,j}}{\sum_{v}A_{k,v}}$ \newline
            }
	\If {$(i,k) \& (j,k) \in \mathcal{E}$ }{
             ${SUM}_{out} \gets {SUM}_{out}+\frac{A_{i,k} A_{j,k}}{\sum_{v} A_{v,k}} $ \newline
            }
	    }
	    $A_{S_{in}}(i,j) \gets {SUM}_{in}$ \newline
	    $A_{S_{out}}(i,j)\gets {SUM}_{out}$ \newline
    }
} 
\underline{Directed Graph Convolution Networks}\newline
$\mathbf{D_F} \gets \text{RowNorm}(A_F)$  \newline
$\mathbf{D_{S_{in}}} \gets \text{RowNorm}(A_{S_{in}})$  \newline
$\mathbf{D_{S_{out}}} \gets \text{RowNorm}(A_{S_{out}})$  \newline  
$\mathbf{Z}_F \gets  \mathbf{\tilde{D}_F^{-\frac{1}{2}}} ~~\mathbf{\tilde{A}_F} ~~\mathbf{\tilde{D}_F}^{-\frac{1}{2}} \mathbf{X} \Theta$\newline 
$\mathbf{Z}_{S_{in}} \gets  \mathbf{\tilde{D}_{S_{in}}}^{-\frac{1}{2}} ~~\mathbf{\tilde{A}_{S_{in}}} ~~\mathbf{\tilde{D}_{S_{in}}}^{-\frac{1}{2}} \mathbf{X} \Theta $\newline
$\mathbf{Z}_{S_{out}} \gets   \mathbf{\tilde{D}_{S_{out}}}^{-\frac{1}{2}} ~~\mathbf{\tilde{A}_{S_{out}}}~~ \mathbf{\tilde{D}_{S_{out}}}^{-\frac{1}{2}} X \Theta $\newline
$\mathbf{Z} = \sigma(\text{Concat}( \mathbf{Z}_F , \alpha \mathbf{Z}_{S_{in}} , \beta  \mathbf{Z}_{S_{out}}))$\newline
$\mathbf{\hat{Y}} = softmax(FCLayer(\mathbf{Z}))$
\end{algorithm}

\subsection{Datasets Details}

We use seven open access datasets to test our method. The origin \textsc{Cora-ML} has 70 classes, we combine the 2 classes that cannot perform the dataset split to the nearest class. 

\begin{table}[h]
\caption{Datasets Details}
\label{dataset}
\small
\begin{tabular}{|c|c|c|c|c|c|c|c|}
\hline
\multirow{2}{*}{\textbf{\emph{Datasets}}}  & \multirow{2}{*}{Nodes} & \multirow{2}{*}{Edges} & \multirow{2}{*}{Classes}  & \multirow{2}{*}{Features} & \multirow{2}{*}{Label rate}\\
   &  &  &  & &\\ \hline
 \hline
 
\textsc{Cora-Full} & 19793 & 65311 & 68\tnote{*} & 8710 &7.07\%\\\hline
\textsc{Cora-ML} & 2995 & 8416 & 7 & 2879 &4.67\%\\\hline
\textsc{CiteSeer} & 3312 & 4715 & 6 & 3703 &3.62\%\\\hline
\textsc{DBLP} & 17716 & 105734 & 4 & 1639 &0.45\%\\\hline
\textsc{PubMed} & 18230 & 79612 & 3 & 500 &0.33\%\\\hline
\textsc{Amazon-Photo} & 7650 & 143663 & 8 & 745 &2.10\%\\\hline
\textsc{-Computer} & 13752 & 287209 & 10 & 767 &1.45\%\\ \hline
 
\end{tabular}
\end{table}

Label rate is the fraction of nodes in the training set per class. We use 20 labeled nodes per class to calculate the label rate.

\subsection{Baselines Details and Settings}

The baseline methods are given below:

\textbullet\ \textbf{ChebNet}: It redefines graph convolution using Chebyshev polynomials to remove the time-consuming Laplacian eigenvalue decomposition.

\textbullet\ \textbf{GCN}: It has multi-layers which stacks first-order Chebyshev polynomials as graph convolution layer and learns graph representations use a nonlinear activation function.

\textbullet\ \textbf{SGC}: It removes nonlinear layers and collapse weight matrices to reduce computational consumption.

\textbullet\ \textbf{GraphSage}: It proposes a general inductive framework that can efficiently generate node embeddings for previously unseen data

\textbullet\ \textbf{GAT}: It applies attention mechanism to assign different weights to different neighborhood nodes based on node features.

For all baseline models, we use their model structure in the original papers, which including layer number, activation function selection, normalization and regularization selection, etc. It is worth noting that GraphSage has three variants in the original article using different aggregators: \textbf{mean}, \textbf{meanpool} and \textbf{maxpool}. In this paper, we use \textbf{mean} as its aggregator as it performs best \cite{shchur2018pitfalls}. Besides, we set the mini-batch size to 512 on \textsc{Amazon-Photo} and \textsc{Amazon-Computer} and 16 on other datasets. For GCN, we set the size of hidden layer to 64 on \textsc{Amazon-Photo} and \textsc{Amazon-Computer} and 16 on other datasets. We also fix the number of attention heads to 8 for GAT, power number to 2 for SGC and $k=2$ in ChebNets, as proposed in the respective papers.


\end{document}